\documentclass[11pt]{article} 

\usepackage[utf8]{inputenc} 

\usepackage{authblk}

\usepackage{geometry} 
\usepackage{graphicx} 

\usepackage{amssymb}
\usepackage{amsmath}
\usepackage{amsthm}
\usepackage{mathtools}
\usepackage{stmaryrd}
\usepackage{xfrac}

\usepackage{booktabs} 
\usepackage{array} 
\usepackage{paralist} 
\usepackage{verbatim} 
\usepackage{subfig} 

\usepackage{fancyhdr} 
\usepackage{sectsty}
\allsectionsfont{\sffamily\mdseries\upshape} 

\usepackage[nottoc,notlof,notlot]{tocbibind} 
\usepackage[titles,subfigure]{tocloft} 

\title{T- Hop: Tensor representation of paths in graph convolutional networks}
\author{Abdulrahman Ibraheem}
\affil[]{rahman.ibraheem1@outlook.com}

\begin{document}
\maketitle

\section{Introduction}

This draft describes a method for capturing path information in graphs, for use in graph convolutional networks (GCN). Let $G= (V, E)$ represent a graph, where $V = \{v_1,...,v_n\}$ and $E = \{e_1, ..., e_m\}$ are the nodes and edges of $G$, as usual. The adjacency matrix of $G$ is $A$. Further, $A^L$ denotes the powered adjacency matrix of $G$, while $A^L_{i,j}$  represents the entry on the $i$-th row and $j$-th column of $A^L$.  The entry $A^L_{i,j}$ corresponds to the number of paths of length $L$ between node $v_i$ and node $v_j$ in $G$. While the seminal \textit{vanilla GCN} of Kipf and Welling \cite{kipf_welling} employs only matrix $A$ as its operator, more recent variants of the GCN, such as MixHop \cite{mix_hop} and Variable Power Networks (VPN) \cite{power_up} employ $A^L$, in their models. The hope of this draft is to improve on the aforementioned methods, by incorporating more information into $A^L$, in hopes that the additional information might lead to higher accuracies in downstream tasks utilizing the resulting representation. Towards this, let us consider two arbitrary nodes, $v_i$ and $v_j $ in graph $G$. Let $\mathcal{B}^{L}_{i,j,k}$  be the number of paths of length $L$ between $v_i$ and $v_j$ that contain $v_k$. Clearly, we can arrange the values, $\mathcal{B}^{L}_{i,j,k}$, in an $n \times n \times n$ 3-d tensor denoted $\mathcal{B}^{L}$, such that the entry on the $i$-th row, $j$-th column and $k$-th depth of $\mathcal{B}^{L}$ is  $\mathcal{B}^{L}_{i,j,k}$. Using  $\mathcal{B}^{L}_{i,j,k}$, we now define a 3-d tensor, $\mathcal{T}^{L}$.   Where $\mathcal{T}^{L}_{i,j,k}$ is the entry on the $i$-th row, $j$-th column and $k$-th depth of $\mathcal{T}^{L}$, I define $\mathcal{T}^{L}_{i,j,k}$ as follows:

\begin{equation}
\mathcal{T}^{L}_{i,j,k} = \frac{ \mathcal{B}^{L}_{i,j,k} } {(L+1)}
\end{equation}
 
From the above definition,  it should be noted that $\mathcal{T}^{L}$ is an $n \times n \times n$ tensor, where $n$ is the number of nodes in the graph. Clearly, working with such a large matrix is computationally demanding. To enhance tractability, I propose applying dimensionality reduction along the depth axis of $\mathcal{T}^{L}$. To underpin the idea, let us fix the row and column indices (i.e $i$ and $j$) of $\mathcal{T}^{L}$, while leaving the depth index to vary. For each pair of indices, $(i, j)$, this results in an $n$-dimensional vector which stretches along the depth axis of $\mathcal{T}^{L}$. Let us denote this vector as $ t^L_{ij} = \mathcal{T}^{L}_{i,j,:}$ . We would like to compute a dimensionality reduction map, $f :\mathbb{R}^n \rightarrow \mathbb{R}^d $, with $d \ll n$, that sends each $t_{ij}^L$ to a $d$-dimensional space. If we apply this map to all $n$-dimensional vectors corresponding to all  $(i,j)$ positions in $\mathcal{T}^{L}$, we get a $n \times n \times d$ tensor, which we denote $\widehat{\mathcal{T}}^{L}$ .  

Indeed, it is quite interesting to observe that if we choose the dimensionality reduction map to be the summation operation, $f_{sum}:\mathbb{R}^n \rightarrow \mathbb{R}$, which simply outputs the sum of all components of  its input vector, then we would obtain the powered adjacency matrix, $A^L$, when we apply $f_{sum}$ to $\mathcal{T}^{L}$. I elaborate on this formally below, beginning with the following definition:

\newtheorem{def1}{Definition}[section] 
\begin{def1}[Cardinality of multiset $\mathcal{P}^L$ ]
\label{def_1}
Let $G$ be a graph of nodes, $V = \{v_1,..., v_n\}$, and edges. Let  $v_i$ and $v_j$ be any two arbitrary nodes in $G$, and Let $A^L_{ij}$ be the number of simple paths of length $L$ between $v_i$ and $v_j$. Let $ P_q^L = \{ v_1^q, v_2^q,..., v_{L+1}^q\}$ denote the $q$-th simple path of  length $L$  between $v_i$ and $v_j$, where $ v_k^q$ is the $k$-th node in the $q$-th path, $P_q^L$. Let  $\mathcal{P}^L = \{P_1^L, P_2^L, ... P_{ A_{ij}}^L \} = \{ v_1^1, v_2^1, ..., v_{L+1}^1,  \;\; v_1^2, v_2^2, ..., v_{L+1}^2, \;\; ..., \;\; v_1 ^{A_{ij}}, v_2 ^{A_{ij}}, ..., v_{L+1} ^{A_{ij}} \} $ be a \textbf{multiset} containing all the simple paths of  length $L$  between $v_i$ and $v_j$. \textbf{Then, the cardinality of multiset $\mathcal{P}^L$  is defined as the number of nodes in $\mathcal{P}^L$, counting multiplicities of nodes. }
\end{def1}

\noindent Based on the preceding definition, the following is a fact:

\newtheorem{thm1}[def1]{Fact }
\begin{thm1}[Cardinality of multiset $\mathcal{P}^L$  equals $\sum_k \mathcal{B}^{L}_{i,j,k}$ ]
\label{thm_1}
 \textbf{The cardinality of multiset $\mathcal{P}^L$ defined in Definition \ref{def_1} above is equal to $\sum_k \mathcal{B}^{L}_{i,j,k}$}
\end{thm1}

\begin{figure}
\includegraphics[height = 120pt, width =200pt]{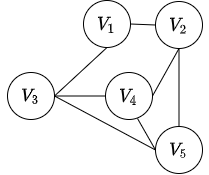} 
\caption{An illustrational graph of five nodes }
\label{fig_1}
\end{figure}

\begin{proof}
 The proof is best sketched with an example. As an example, let us use the graph $G$ of five nodes depicted in Figure \ref{fig_1}. Without loss of generality, let us consider all simple paths of length $L = 3$ between two arbitrary nodes, $v_1$ and $v_5$ in the graph. From the graph, we see that there are 2 simple paths of length $3$ between $v_1$ and $v_5$, so that $A_{1,5}^3 = 2$. These paths are $P_1^3 = \{v_1, v_2, v_4, v_5\}$ and  $P_2^3 = \{v_1, v_3, v_4, v_5\}$. Hence, we may write 
$\mathcal{P}^3 = \{P_1^3, P_2^3\} = \{v_1, v_2, v_4, v_5 \;\; v_1, v_3, v_4, v_5\}$. Upon sorting $\mathcal{P}^3$, we now have $\mathcal{P}^3 = \{v_1, v_1,  \;\; v_2,\;\; v_3, \;\; v_4, v_4,  \;\; v_5, v_5 \}$. Now, given any node, $v_k$ in $G$, when we count the multiplicity of $v_k$ in the sorted version of $\mathcal{P}^3$, we see it corresponds to the number of simple paths of length $3$ between $v_1$ and $v_5$ that contain $v_k$. For example, we see clearly that node $v_4$ has multiplicity of 2, because it is contained in two different paths of length 3 between $v_1$ and $v_4$, whereas node $v_2$ has multiplicity of $1$, because it is contained in a single path of length 3 between $v_1$ and $v_5$. Generalizing this observation, we see that, if $\mathcal{P}^L$ is the multiset of nodes that constitute the simple paths of length $L$ between any two arbitrary nodes, $v_i$ and $v_j$, then for any  node, $v_k \in \mathcal{P}^L$ , the multiplicity of $v_k$ in $\mathcal{P}^L$ corresponds to the number of paths of length $L$ between $v_i$ and $v_j$ that contain $v_k$, which in turn, by definition, is equal to  $\mathcal{B}^{L}_{i,j,k}$. \textbf{In summary, for any  node, $v_k \in \mathcal{P}^L$, the multiplicity of $v_k$ in $\mathcal{P}^L$ equals $\mathcal{B}^{L}_{i,j,k}$}.  Based on this, we now consider the quantity $\sum_k \mathcal{B}^{L}_{i,j,k}$. It should be clear that the quantity $\sum_k \mathcal{B}^{L}_{i,j,k}$ simply equals the sum of multiplicities of all nodes in  $\mathcal{P}^L$, which, in turn, equals the cardinality of $\mathcal{P}^L$.
\end{proof}

\noindent Next, using  Fact \ref{thm_1}, we have the following proposition: 

\newtheorem{thm2}[def1]{Proposition } 
\begin{thm2}[$f_{sum}$ recovers $A^L$ from  $\mathcal{T}^{L}$  ]
Let $f_{sum}:\mathbb{R}^n \rightarrow \mathbb{R}$ be the function that takes a vector $u \in \mathbb{R}^n$ as input and returns as output the summation of all components of $u$. Then, with  $ t^L_{ij}$ denoting the $n$-dimensional vector that stretches along the depth-axis of the $3$-d tensor, $\mathcal{T}^{L}$, at a given $i$-th row, $j$-th column position of $\mathcal{T}^{L}$, we have that $f_{sum}(t_{ij}^L) = A^L_{ij}$. 
\end{thm2}
\begin{proof}

To proceed, from Definition \ref{def_1} above, we recall  the meaning of the multiset $\mathcal{P}^L = \{P_1^L, P_2^L, ... P_{ A_{ij}}^L\}   =  \{ v_1^1, v_2^1, ..., v_{L+1}^1,  \;\; v_1^2, v_2^2, ..., v_{L+1}^2, \;\; ..., \;\; v_1 ^{A_{ij}}, v_2 ^{A_{ij}}, ..., v_{L+1} ^{A_{ij}} \}$; we also bring to mind the definition of the multiset's cardinality given therein. In particular, the number of paths in $\mathcal{P}^L$ is $A_{ij}$ and each path contains  $L + 1$ nodes, so that the cardinality of $\mathcal{P}^L$ is equal to  $(L+1)A_{ij}^L$.  Hence, we have $|\mathcal{P}^L| = (L+1)A_{ij}^L$. But, we already know from  Fact \ref{thm_1} above that  $ |\mathcal{P}^L|  = \sum_k \mathcal{B}^{L}_{i,j,k}$.  Hence, we have $ \sum_k \mathcal{B}^{L}_{i,j,k} = (L+1)A_{ij}^L$, implying $ \sum_k \dfrac{ \mathcal{B}^{L}_{i,j,k} }{(L+1)} = A_{ij}^L $. Further, by definition, we know $\dfrac{ \mathcal{B}^{L}_{i,j,k} }{(L+1)} = \mathcal{T}^{L}_{i,j,k}$. Thus, $\sum_k \mathcal{T}^{L}_{i,j,k} = A_{ij}^L$. Now, it is clear that $ \sum_k \mathcal{T}^{L}_{i,j,k}$ is tantamount to applying $f_{sum}$ to  the vector $  t^L_{ij} = \mathcal{T}^{L}_{i,j,k}$, which completes the proof.
\end{proof}

In the above, the application of $f_{sum}$ to the a $n \times n \times n$ 3-d tensor, $\mathcal{T}^{L}$, to obtain the $n \times n $ 2-d matrix, $A^L$, can be viewed as a dimensionality reduction process. However, it seems that applying the summation operation, embodied in $f_{sum}$, to  $\mathcal{T}^{L}$ might result in too much loss of the information encoded in  $\mathcal{T}^{L}$. It is only natural to hope that a \textit{softer} approach which compresses the original information contained in $\mathcal{T}^{L}$ into  a $n \times n \times d$ tensor, with $d \ll n$, might result in lesser loss of information, and might thereby lead to improved accuracies in downstream tasks. This is why a key proposal of this draft is to explore the use of dimensionality techniques to compress the information encoded in  $\mathcal{T}^{L}$ into a $n \times n \times d$ tensor, namely $\widehat{\mathcal{T}}^{L}$. For dimensionality reduction, one could explore the variational autoencoder and its variants \cite{auto_encoder} \cite{vae}. 

 After dimensionality reduction, we take the resulting  tensor, $\widehat{\mathcal{T}}^{L}$, and employ it within a larger GCN framework such as MixHop. In MixHop,  the $(l + 1)$-th layer of GCN performs the following operation:

\begin{equation}
\label{eqn_1}
H^{l+1} =  \bigparallel_{L \in P} \sigma( A^L H^{l}W_{L}^{l})
\end{equation}

In the preceding equation, $P$ is an index set, which acts as an hyperparameter of the model. For example, one could have $P = \{ 0, 1, 2 \}$.  Further,  $H^{l} \in \mathbb{R}^{n \times s_l} $ represents the input to the $l$-th GCN layer, with $n$ being the number of nodes in the underlying graph. $A^L  \in \mathbb{R}^{n \times n}$ denotes the $L$-power adjacency matrix, $W_{L}^l \in \mathbb{R}^{s_{l} \times \widehat{s}_{l+1}}$ is a learnable parameter of the model, $\sigma(.)$ denotes an activation function, and $H^{l+1} \in \mathbb{R}^{n \times s_{l+1}}$ denotes the output from the $l$-th GCN layer. Also, $\bigparallel$ denotes concatenation of $\sigma( A^L H^{l}W_{L}^{l})  \in \mathbb{R}^{n \times \widehat{s}_{l+1}}$  along the column dimension. Due to this concatenation operation, we have $s_{l+1} =  |P| \widehat{s}_{l+1}$, where $|P|$ is the cardinality of set $P$.

I now turn to define my proposed \textbf{Tensor Hop} (T-Hop) model. To proceed, let $\widehat{\mathcal{T}}^{L}_{:,:,k}$ denote the $n \times n$  matrix obtained at the $k$-depth of $\widehat{\mathcal{T}}^{L}$. My proposal is to use $\widehat{\mathcal{T}}^{L}_{:,:,k}$ in place of $A^L$ in the MixHop layer of Equation \ref{eqn_1} above. To formalize this, I define my proposed T-Hop layer as:

\begin{equation}
\label{eqn_2}
H^{l+1} =  \bigparallel_{L \in P}  \bigoplus_{k =1}^{d} \; \sigma( \widehat{\mathcal{T}}^{L}_{:,:,k} \: H^{l}W_{L}^{l} )
\end{equation}

The notation in Equation \ref{eqn_2} above is similar to that in Equation \ref{eqn_1}, except that $\bigoplus$ denotes a generic aggregation operation and $d$ denotes the number of depth in the $n \times n \times d$ tensor, $\widehat{\mathcal{T}}^{L}$ . An example aggregation operation that could be used for implementing  $\bigoplus$ is the element-wise averaging operation of two or more matrices. Comparing the proposed T-Hop layer in Equation \ref{eqn_2} with the MixHop layer of Equation \ref{eqn_1}, we see that there is a very high degree of correspondence between the two of them. In particular, in Equation \ref{eqn_2}, the matrix $\: \widehat{\mathcal{T}}^{L}_{:,:,k} \in \mathbb{R}^{n \times n}$ plays the role which  $A^L  \in \mathbb{R}^{n \times n}$ plays in Equation \ref{eqn_1}.  Moreover, the same set of parameters, $W_{L}^{l}$, are \textbf{shared} for all $k = 0,1,..., d$  in the computation of $\sigma( \widehat{\mathcal{T}}^{L}_{:,:,k} \: H^{l}W_{L}^{l} )$ in  Equation \ref{eqn_2}.  Notice that, as long as $d \ll n$ is satisfied, this parameter sharing scheme makes the computational complexity of the proposed T-Hop layer to be on the order of that of the MixHop layer. Moreover, the memory footprint of the proposed T-Hop layer is also on the order of that required for MixHop, as long as $d \ll n$ is satisfied.

\end{document}